\documentclass[a4paper, 10pt, conference]{ieeeconf}      
\usepackage{FG2020}

\FGfinalcopy 

\IEEEoverridecommandlockouts                              
\usepackage{graphics} 
\usepackage{epsfig} 
\usepackage{mathptmx} 
\usepackage{times} 
\usepackage{amsmath} 
\usepackage{amssymb}  
\usepackage{subfigure}
\usepackage{url}
\usepackage{color}

\def\FGPaperID{167} 

\title{\LARGE \bf
Mutual Information Maximization for Effective Lip Reading
}

\author{\parbox{16cm}{\centering
    {\large Xing Zhao${^1}^*$\thanks{$^*$ This work is done during Xing Zhao's internship at Institute of Computing Technology, Chinese Academy of Sciences.}, Shuang Yang$^2$, Shiguang Shan$^{2,3}$, Xilin Chen$^{2,3}$}\\
    {\normalsize
    $^1$ Zhejiang University of Technology, Hangzhou 310014, China\\
    $^2$ Key Laboratory of Intelligent Information Processing of Chinese Academy of Sciences (CAS), Institute of Computing Technology, CAS, Beijing 100190, China\\
    $^3$ University of Chinese Academy of Sciences, Beijing 100049, China}}
}

\begin{document}

\ifFGfinal
\thispagestyle{empty}
\pagestyle{empty}
\else
\author{Anonymous FG2020 submission\\ Paper ID \FGPaperID \\}
\pagestyle{plain}
\fi
\maketitle

\begin{abstract}
Lip reading has received an increasing research interest in recent years due to the rapid development of deep learning and its widespread potential applications. One key point to obtain good performance for the lip reading task depends heavily on how effective the representation can be to capture the lip movement information and meanwhile to resist the noises resulted from the change of pose, lighting conditions, speaker's appearance, speaking speed and so on. Towards this target, we propose to introduce the mutual information constraints on both the local feature's level and the global sequence's level to enhance the relations of the features with the speech content. On the one hand, we constraint the features generated at each time step to enable them carry a strong relation with the speech content by imposing the local mutual information maximization constraint (LMIM), leading to improvements over the model's ability to discover fine-grained lip movements and the fine-grained differences among words with similar pronunciation, such as ``spend'' and ``spending''. On the other hand, we introduce the mutual information maximization constraint on the global sequence's level (GMIM), to make the model be able to pay more attention to discriminate key frames related with the speech content, and less to various noises appeared in the speaking process. By combining these two advantages together, the proposed method is expected to be both discriminative and robust for effective lip reading. To verify this method, we evaluate on two large-scale benchmarks whose videos are all collected from different TV shows, covering a wide range of various speaking conditions. We perform a detailed analysis and comparison on several aspects, including the comparison of the LMIM and GMIM with the baseline, the visualization of the learned representation and so on. The results not only prove the effectiveness of the proposed method but also report new state-of-the-art performance on both the two benchmarks.
\end{abstract}

\section{Introduction}
Lip reading is a task to infer the speech content in a video by using only the visual information, especially the lip movements. It has many crucial applications in practice, such as assisting audio-based speech recognition \cite{chen1998audio}, biometric authentication \cite{assael2016lipnet}, aiding hearing-impaired people \cite{xu2018lcanet}, and so on. With the huge success of deep learning based models for several related tasks in the computer vision domain, some works began to introduce the powerful deep models for effective lip reading in these years \cite{assael2016lipnet,stafylakis2017combining,stafylakis2018pushing,petridis2018end}. For example, \cite{stafylakis2017combining} proposed an end-to-end deep learning architecture for word level visual speech recognition, which is a combination of a convolutional network with a  bidirectional Long Short-Term Memory network, yielding an improvement of 6.8\% on the accuracy than before. Besides the great impetus of deep learning technologies, several large-scale lip reading datasets, were released in recent years, such as LRW \cite{chung2016lip}, LRW-1000 \cite{yang2018lrw}, LRS2 \cite{chung2017lip}, LRS3 \cite{afouras2018lrs3}, and so on. These datasets have also contributed significantly to the recent progress of lip reading.
\begin{figure}[t]
\centering
\subfigure[An example of a video sample with the annotated label ``ABOUT'']{\label{fig1_a}\includegraphics[width=1.0\linewidth]{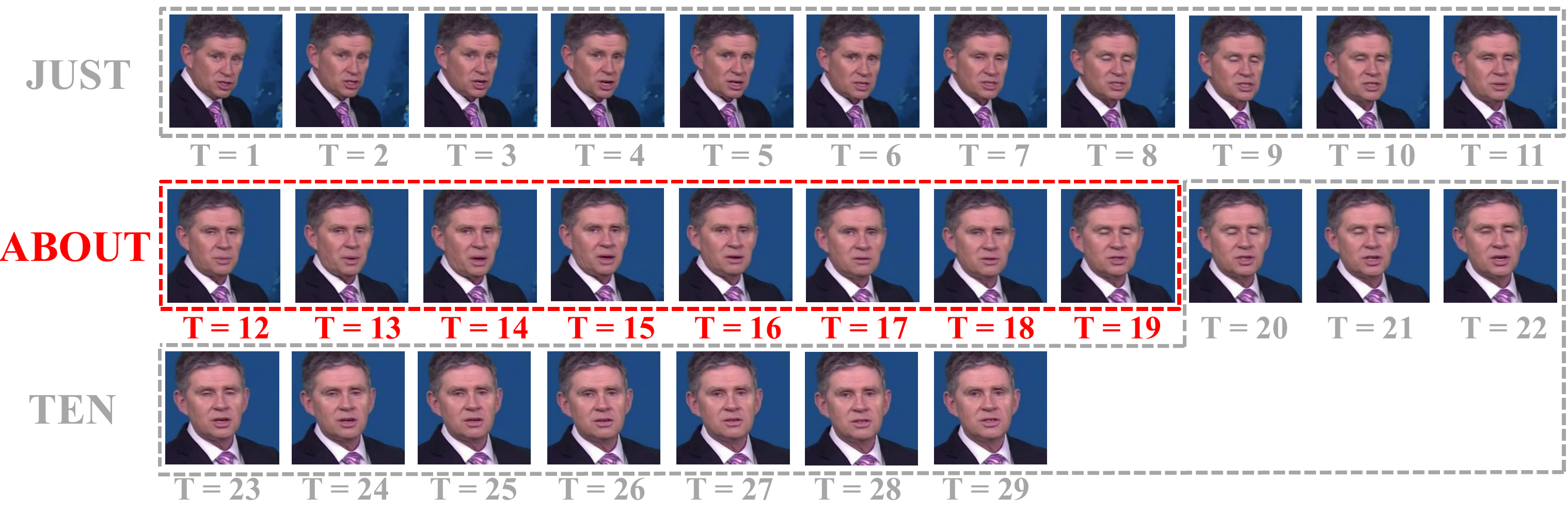}}
\subfigure[Another two samples of ``ABOUT'']{\label{fig1_b}\includegraphics[width=1.0\linewidth]{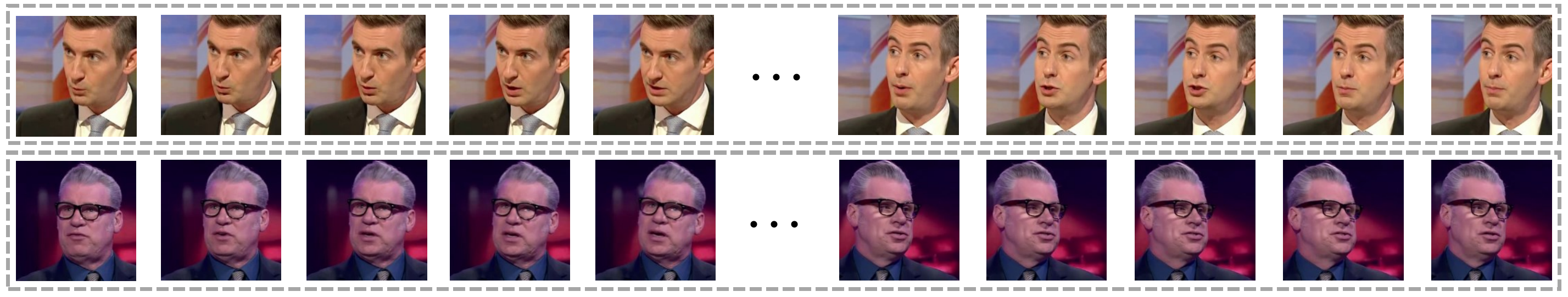}}
\caption {The word-level lip reading is a challenge task. \textbf{(a)} The actual frames of the annotated word ``ABOUT'' include only frames at the time step T = 12$\sim$19. \textbf{(b)} The same word label always have a greatly diversified appearances changes.}
\label{new_fig1} 
\end{figure}

In this paper, we focus on the word-level lip reading, which is a basic but important branch in the lip reading domain. For this task, each input video is annotated with a single word label even when there are other words in the same video, as shown in Fig. \ref{new_fig1}. For example, the video sample in Fig. \ref{fig1_a}, including 29 frames in total, is annotated as ``ABOUT'', but the actual frames of the word ``ABOUT'' include only frames at time step T = 12$\sim$19, shown in the red boxes. The frames before and after this interval are corresponding to the word ``JUST'' and ``TEN'' respectively, not ``ABOUT''. This is consistent with the actual case where the exact boundary of a single word is always hard to get. This property requires a good lip reading model to be able to learn the latent but consistent patterns reflected in different videos with the same word label, and so able to pay more attention to valid key frames, but less to other unrelated frames. Besides the challenges of inaccurate word boundaries, the video samples corresponding to the same word label always have greatly diversified appearance changes, as shown in Fig. \ref{fig1_b}. All these properties require the lip reading model to be able to resist the noises in the sequence to capture the consistent latent patterns in various speech conditions.

In the meanwhile, due to the limited effective area of lip movements, different words probably show similar appearance in the speaking process. Especially, the existence of homophones where different words may look the same or quite similar increases many extra difficulties to this task. These properties require the model being able to discover the fine-grained differences related to different words in the frame-level to distinguish each word from the other.

To solve the above issues, we introduce the mutual information maximization (MIM) on different levels to help the model learn both robust and discriminative representations for effective lip reading. On the one hand, the representations at the global sequence level would be required to have a maximized mutual information with the speech content, to force the model learning the latent consistent global patterns of the same word label in different samples, while being robust to the variations of pose, light and other label-unrelated conditions. On the other hand, the features at the local frame level would be required to maximize their mutual information with the speech content to enhance the word-related fine-grained movements at each time step to further enhance the differences between different words. By combining these two types of constraints together, the model could automatically find and distinguish the valid important frames for the target word, and ignore other unrelated frames. Finally, we evaluate the proposed approach on two large-scale benchmarks LRW and LRW-1000, whose samples are all collected from various TV shows with a wide variation of the speaking conditions. The results show a new state-of-the-art performance on both the two challenging datasets when compared with other related work in the same condition of using no extra data or extra pre-trained models. 

The proposed method could also be easily modified to other existing models for other tasks, which may bring some meaningful insights to the community for other tasks.
   \begin{figure*}

      \centering
      \includegraphics[width=0.9\linewidth]{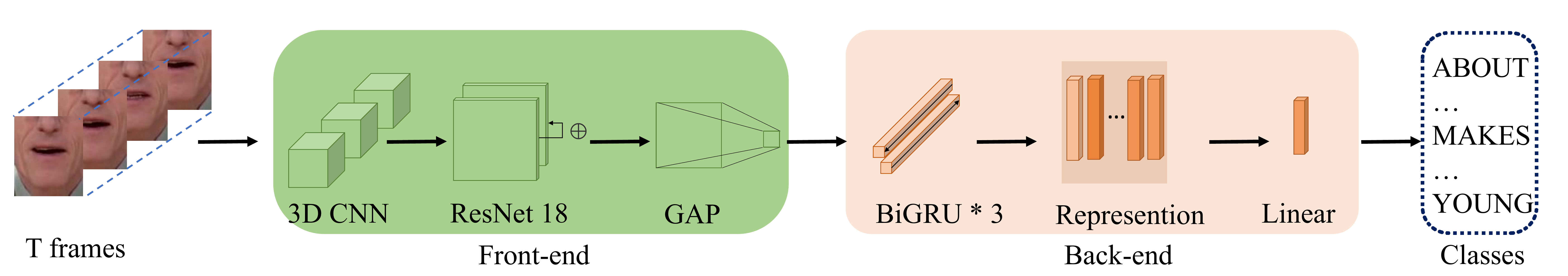}
      \caption{The base architecture.}
    \label{fig:1}
   \end{figure*}
\section{Related Work}
In this section, we provide an overview of the related literature on two closely related aspects, lip reading and mutual information based methods.
\subsection{Lip Reading}
When deep learning technologies are not so popular, many methods have achieved several encouraging results by using specifically-designed and hand-engineered features, such as optical flow \cite{shaikh2010lip}, lip landmarks tracking \cite{duchnowski1995toward}, and so on. The classification is often done by Support Vector Machine \cite{shaikh2010lip} together with the Hidden Markov Models (HMMs) \cite{chandrasekaran2009natural,papandreou2009adaptive}. We refer to \cite{zhou2014review,potamianos2004audio} for a detailed review on these non-deep methods for lip reading. These previous work have provided an important impetus to the advancement of lip reading at the early stage.

With the rapid development of deep learning in recent years, more and more researchers gradually tend to perform the lip reading task by deep neural networks. 

2D-CNN is the first type of network applied to lip reading to extract features for each frame. \cite{noda2014lipreading} proposed a system including a CNN and a hidden Markov model with Gaussian mixture observation model (GMM-HMM). The outputs of the CNN are regarded as visual feature sequences, and the GMM-HMM is applied on this sequence for word classification. In the later works \cite{wand2016lipreading,chung2017lip}, long short-term memory (LSTM) or gated recurrent unit (GRU) is used to model the patterns in the temporal dimension. The CNN-LSTM based models, which can be trained in an end-to-end manner, has gradually become a processing pipeline for lip reading.

However, the mouth regions in different frames are not always aligned at exactly the same position. So the context shown in nearby frames always plays an important role for effective lip reading. Several methods introduce the 3D convolution operation to tackle this problem \cite{petridis2018end,stafylakis2018pushing,yang2018lrw}. For example, LipNet \cite{assael2016lipnet} employed a 3D-CNN at the front-end on the visual frames and obtained remarkable performance for lip reading. Stafylakis et al. \cite{stafylakis2017combining} combined a 3D-CNN and a 2D-CNN based network to obtain robust features, which got a much higher accuracy on LRW dataset than before.

Besides directly applying different types of deep networks to lip reading, some recent impressive works begun to design particular modules to solve the shortcomings of some existing networks for more effective lip reading. For example, Stafylakis et al. \cite{stafylakis2018pushing} introduced additional word boundary information to improve the performance on the word-level LRW dataset. \cite{chung2017lip} employed the attention mechanism to select key frames in a sequence-to-sequence model. Wand et al. \cite{wand2017improving} improved the accuracy of lip reading by domain-adversarial training, which is expected to get speaker-independent features, beneficial to the final word classification. However, their method is hard to apply when coming to a large scale dataset with large number of speakers. Recently, Wang \cite{wang2019multi} extracted both frame-level fine-grained features and short-term medium-grained features by a 2D-CNN network and 3D-CNN network respectively. In this paper, we propose a new way for effective lip reading. Specifically, we introduce the constraints on both the local feature level and the global representation level to make the model both be able to learn fine-grained features and pay attention to key frames respectively. 
\subsection{Mutual Information Mechanism}
Mutual information (MI) is a fundamental quantity for measuring the relationship between two random variables. It is always used to evaluate the ``amount of information'' owned by one random variable when given the other random variable. Based on this property, the mutual information of two random variables is always used as a measure of the mutual dependence between two variables. Moreover, unlike the Pearson correlation coefficient which only captures the information in the degree of linear relationship, mutual information also captures nonlinear statistical dependencies \cite{kinney2014equitability}, and therefore has a wide range of applications.

For example, Ranjay et al. \cite{krishna2019information} solve the visual question answer problem by maximizing the MI between the image, the expected answer and the generated question, leading to the model’s ability to select corresponding powerful features. Li et al. \cite{li2016mutual} tried to maximize the MI between the source and target sentences in the neural machine translation task to improve the diversity of translation results.

One work which has a bit relation with our work is Zhu et al. \cite{zhu2018high}, who performed talking face generation by maximizing the MI between the words distribution and the facial/audio distribution. But in our work, we try to maximize the MI between the words distribution and the representation at different levels, to guide the model towards learning both robust and discriminative features for the lip reading task, which is totally different with \cite{zhu2018high}.

   \begin{figure*}
      \centering
      \includegraphics[width=0.9\linewidth]{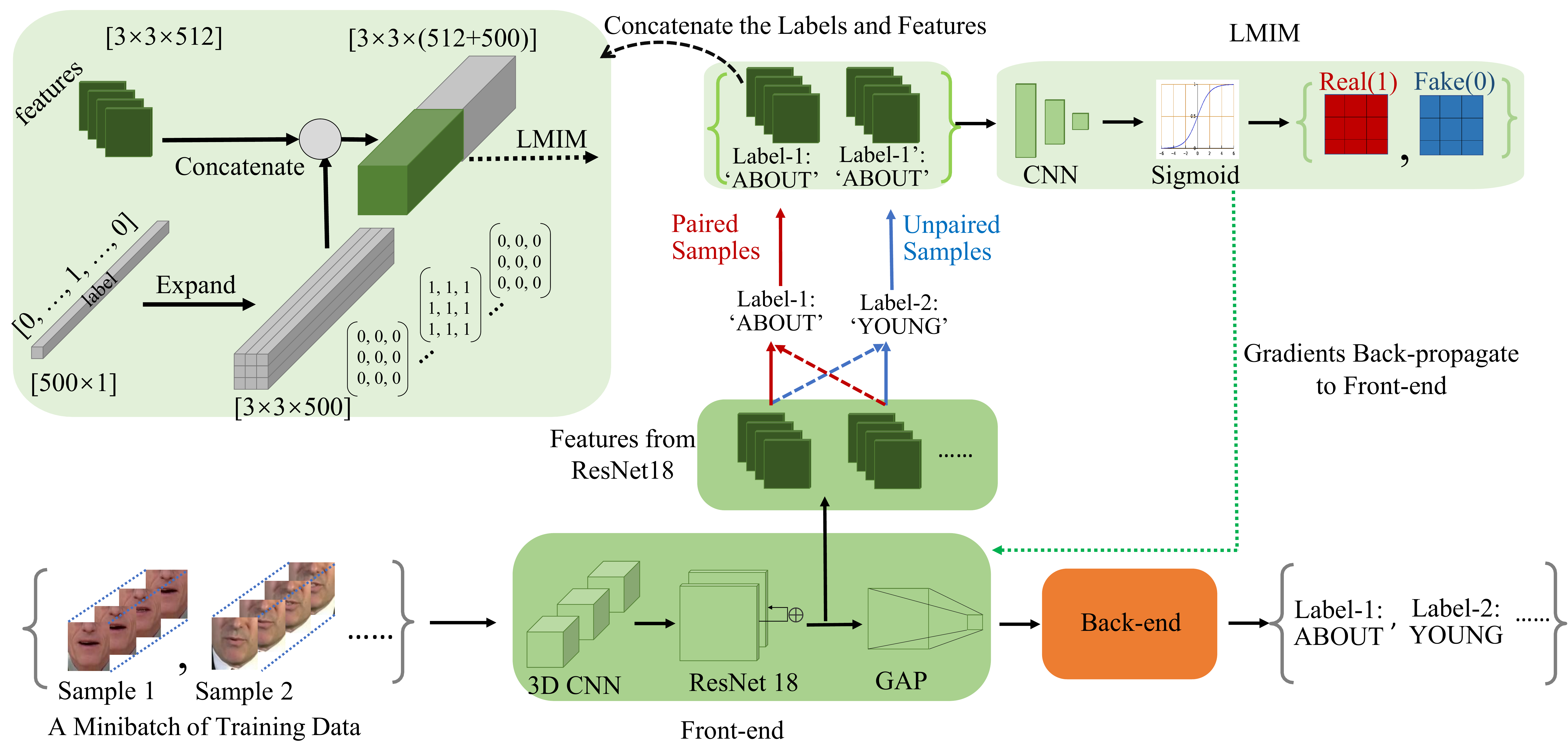}
      \caption{The process of training the base network with the proposed LMIM. The total loss is computed by averaging over all the time steps and patches. The gradients from the LMIM will be back-propagated to the Front-end through the features sampled from the ResNet18. The LMIM will be dropped after training.}
    \label{fig:2}
   \end{figure*}
\section{THE PROPOSED MUTUAL INFORMATION MAXIMIZATION FOR LIP READING}
In this section, we would first give an overview to the overall architecture. Then the particular manner to impose mutual information mechanisms on different levels is presented. Finally, the optimization process to learn the model is provided.
\subsection{The Overall Architecture}\label{3.1}
Let $\textbf{X}=(x_1, x_2, ..., x_T)$ denotes the input sequence with $T$ frames in total, where $x_i$ is the feature vector of the $i$-th frame. The task of the model is to classify the input sequence $\textbf{X}$ into one of the $C$ classes, where $C$ is the number of classes. Let $\textbf{Y}= (0, 0, 1, ..., 0)$ denotes the annotated word label of the sequence, where $\textbf{Y}$ is a $C-$dimensional one-hot vector with only a single 1 at the position corresponding to its word label index. We construct our base architecture with two principal components, named as front-end and back-end respectively, which enable the total network to be trained end-to-end. 

The \textbf{Front-end} includes a 3D-CNN layer, a spatial pooling layer, a ResNet18 network, and a GAP layer, as shown in Fig. \ref{fig:1}. Specifically, given the input image sequence $\textbf{X}$, a 3D-CNN layer is firstly applied on the raw frames, in order to perform an initial spatial temporal alignment in the sequence for effective recognition. A spatial max-pooling layer is then followed to compact the features in the spatial domain. It should be noted that we keep the temporal dimension unchanged in this procedure to avoid a further shortage of the movement information in the sequence because the duration of each word is always very short. In the next step, we divide the features into $T$ parts and employ a ResNet18 module at each time step $t=1, 2, ..., T$ to separately extract discriminative features. To improve the ability to capture fine-grained movements related to the spoken word, we impose the mutual information constraint on the pairs of outputs of ResNet18 and the annotated label. Having been maximized the relations with the annotated label, all these features obtained from the ResNet18 module would be fed into a global average pooling(GAP) layer to compress into $T \times D$-dimensional outputs, where D is the channel of the last layer and 512 in this paper.

With the initial representation from the Front-end, the \textbf{Back-end}, as shown in Fig. \ref{fig:1}, include a 3-layer Bi-GRU network and a linear layer, to capture and classify the latent patterns of the sequence. A Bi-GRU contains two independent single directional GRUs. The input sequence is fed into one GRU in the normal order, and into another GRU in the reverse order. The outputs of the two GRUs would be concatenated together at each time step to represent the whole sequence. The output of the Bi-GRU is expected to be a global representation of the whole input sequence with dimension $T \times 2N$, where $N$ is the number of hidden neurons in each GRU. The representation will be finally sent to a linear layer for classification. To improve its ability to resist noises and select key frames in the sequence, we impose the second mutual information constraint on this global representation. 

\subsection{Local Mutual Information Maximization (LMIM)}
As stated in the previous section, the performance of lip reading is heavily affected by the model’s ability to capture the local fine-grained lip movements, so as to generate discriminative features to distinguish different words from each other. The MI-based constraint is a promising tool for learning good features in an unsupervised way, because we never need any extra data to train it. As stated above, we would introduce Local Mutual Information Maximization (LMIM) on ResNet18 to help the model focus more on related spatial regions at each time step and produce more discriminative features. For lip reading, the local features nearby the mouth regions are significant for the final accurate recognition. Therefore, unlike most existing work \cite{krishna2019information,zhu2018high}, we perform maximization of the MI on each patch of the feature maps rather than the whole feature maps.

Because mutual information is notoriously hard to compute for unknown distribution, we estimate it with the help of deep network here. Following the representation of Jensen-Shannon(JS) MI estimator \cite{hjelm2018learning,nowozin2016f}:
\begin{equation}
\begin{split}
\widehat{\mathcal{I}}_{\theta}^{(J S D)}(A, B) &= \mathbf{E}_{p(A, B)}\left[-\varphi\left(-T_{\theta}(a, b)\right)\right] \\
&- \mathbf{E}_{p(A) p(B)}\left[\varphi\left(T_{\theta}(a, b)\right)\right],
\end{split}
\label{4}
\end{equation}
\noindent where $\varphi(k)=log(1+e^k)$, $A$ and $B$ are the two variables that we want to estimate the MI between them,  $T_{\theta}$ is a continuous function that we directly use a network to approximate it. The ${p(A,B)}$ is the joint distribution of paired samples $\{a,b\}$, and the ${p(A)p(B)}$ is the marginal distribution of the unpaired samples $\{a,b\}$ by randomly sampling ${A}$ and ${B}$. In the optimization process, because $\varphi(k)=log(1+e^k)$ is a monotone increasing function, so maximizing the JS MI estimator is equivalent to optimize (\ref{4}) with $\varphi(k)=log(1+k)$ when the formula is equal to the binary cross-entropy loss. 

With the estimation above, the process of the LMIM is shown in Fig. \ref{fig:2}. We assume the feature map in the last layer of ResNet18 (which will be sent to the GAP layer) as \textbf{F} with a shape of $H \times W \times D$, where $H, W$ and $D$ are the height, width and the channels respectively. Then we divide the feature \textbf{F} into $H \times W$ local patches $(f_1, f_2, ..., f_{H \times W})$ which looks like we separate the original frame to $H \times W$ patches when the receptive field of the features are mapped to the original frame. The label of each sample is expanded by repetition from one-hot vector of dimension $C\times 1$ to the same height and width as $C \times H \times W$. Then we concatenate the labels and features together to obtain a representation of dimension $(C+D) \times H \times W$, which would be used as the input to estimate the Local Mutual Information Maximization network (LMIM). To obtain the local mutual information at each position of the $H \times W$ locations, we employ two convolutional layers with kernel size $1 \times 1$ on the concatenated representation. Then a sigmoid activation is applied to the last layer to simulate the value of the mutual information. Please note that the architecture of the network in this step can be any other form, because it is just applied to approximate a continuous function $T_{\theta}$. But the output layer should always be based on a sigmoid activation function to employ the binary cross-entropy based estimation. The dimension of the outputs of LMIM is $H \times W$, with each number illustrating the degree of how much the corresponding patch is related with the given word label. In the learning process, we expect the mutual information of every patch close to 1 (Real) if the features and the labels are of the same sample (paired samples), and 0 if the label is different with the annotated label of the input sequence (unpaired sample). To collect unpaired samples, we randomly concatenated the features with other labels in the same batch in the implementation process.

Therefore, the optimization for LMIM can be denoted as a binary cross-entropy loss as:
\begin{equation}
\begin{split}
L_{(LMIM)} &= {E}_{p(F, Y)}\left[\log \left(LMIM\left(f, y\right)\right)\right] \\
&+ {E}_{p(F)p(Y)}\left[\log \left(1-LMIM\left(f, y\right)\right)\right].
\end{split}
\end{equation}
\noindent Noting that in this stage, we have not any special process in the temporal dimension. The features of T time steps in an input video will be sent to LMIM successively. In the end, the mean of the loss at all time steps is computed to obtain the gradients for subsequent update.
   \begin{figure*}[t]
   
      \centering
      \includegraphics[width=0.9\linewidth]{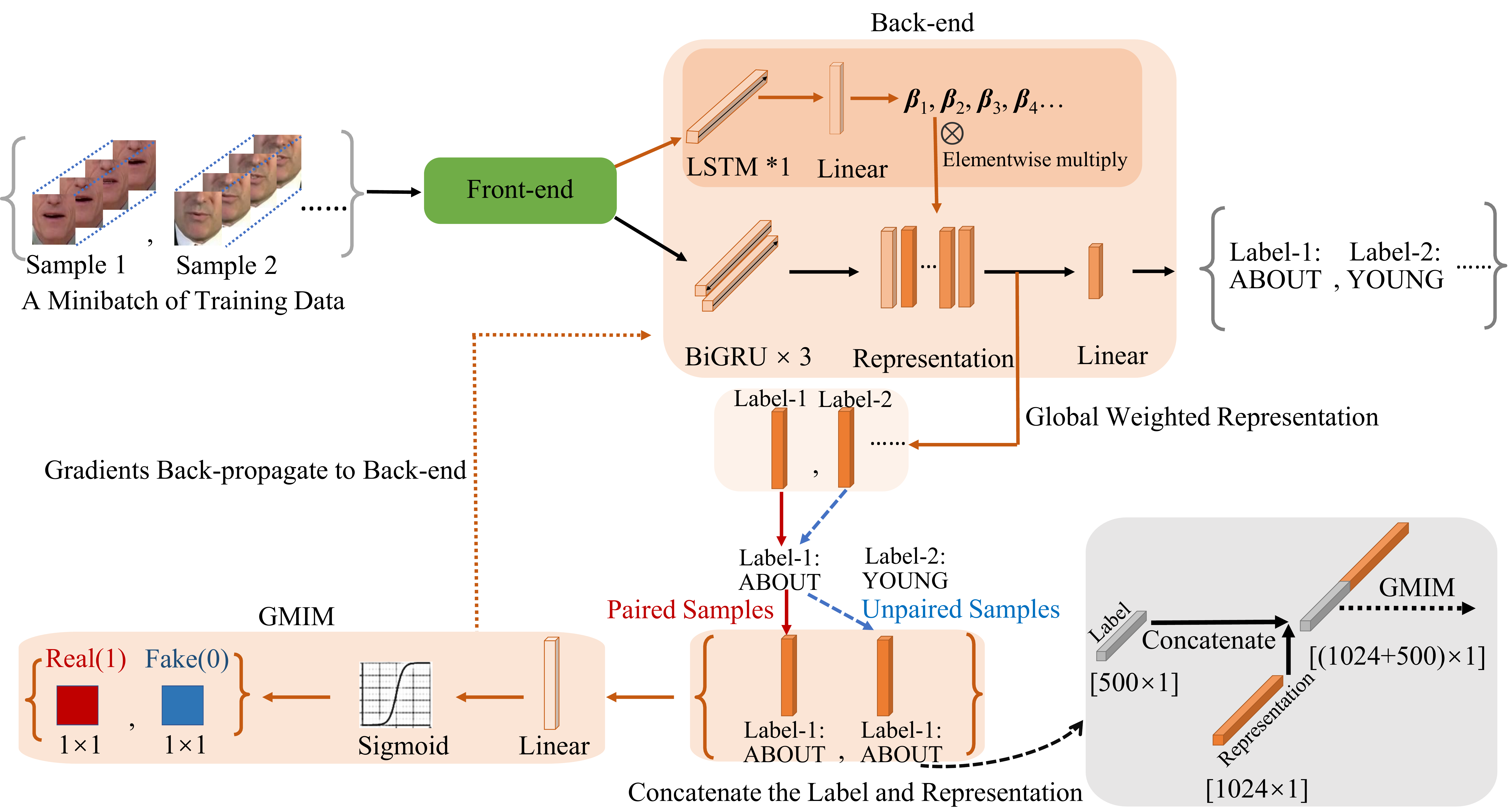}
       \caption{The process of training the network with the proposed GMIM, noted that when we apply the GMIM, a single layer LSTM and a linear layer are also added to the Back-end for computing the weight of each frame, it will be retained after training while the GMIM will be dropped.}
    \label{fig:3}
   \end{figure*}
\subsection{Global Mutual Information Maximization (GMIM)}\label{3.3}
In each sequence, the amount of valuable information provided by different frames is not equal for robust lip reading. In several practical cases, there are many frames corresponding with other words than the given target word in a given sequence. One popular way in current related methods is to average over all the time steps to get the final representation, which would suffer superior performance when coming to practice.

In this paper, we introduce global mutual information maximization on the global representation obtained by the Bi-GRU. Specifically, we introduce an additional LSTM together with a linear layer over the outputs of the Front-end. This additional LSTM would assign different weights $\mathbf{\beta}$ ($T \times 1$-dimensional) for different frames according to the target word. The total architecture is shown in Fig. \ref{fig:3}.

Based on the outputs $\mathbf{Z}$ with dimension $T \times 2N$ of the 3-layer Bi-GRU layers and the weighted value $\mathbf{\beta}$, the final global representation is obtained as the weighted average of the outputs $\mathbf{Z}$ as:
\begin{equation}
\mathbf{O} = \frac{\sum_{t=1}^{T} \beta_{t} \cdot \mathbf{Z}_{t}}{T}.
\end{equation}
\noindent The output $\mathbf{O}$ of dimension $2N$ is then sent to a linear layer to transform its shape from $2N$ to $C$, where $C$ is the number of classes. Specifically, the final representation of the whole sequence $\mathbf{O}'$ of dimension $C$ is applied to get the classification score as
\begin{equation}
\hat{Y}_{i}=\frac{\exp \left(O_{i}\right)}{\sum_{j=1}^{C} \exp \left(O_{j}\right)}.
\end{equation}
\noindent For related valuable key-frames, the weight $\mathbf{\beta}$ should be positive and can be of any value in our method. While for unrelated frames, we just want its weight close to zero, not a negative number for the optimization problem. Therefore we use ReLU to obtain the weight $\beta$ as
\begin{equation}
\beta_{t}=\operatorname{ReLU}\left(\mathbf{W}_{linear} \times \mathbf{LSTM}(\mathbf{G})_{t} + \mathbf{b}_{linear}\right),
\end{equation}
\noindent where $\mathbf{G}$ is the outputs of the GAP layer, $\mathbf{W}_{linear}$ and $\mathbf{b}_{linear}$ are the parameters of the linear layer and  $\mathbf{LSTM}(\mathbf{G})_{t}$ denotes the hidden state at time step $t$ of the extra LSTM layer.

To guide the learning of the weights, we constrain the weighted average vector to contain most of the information about the target word. Specifically, we maximize the MI between the above weighted average representation $\mathbf{O}$ and the annotated label $\textbf{Y}$, both of which will be fed into the global mutual information maximization module (GMIM), which consists of two linear layers and outputs a scalar after a sigmoid activation. Similarly to LMIM, If  $\mathbf{O}$ and $\textbf{Y}$ come from paired samples, we expect the outputs of GMIM as large as possible and even close to 1 (Real). In other cases, the output is expected to be close to 0 (Fake). So the objective function can be written as:
\begin{equation}
\begin{split}
L_{(GMIM)} &= {E}_{p(O, Y)}\left[\log \left(GMIM\left(o, y\right)\right)\right] \\
&+ {E}_{p(O)p(Y)}\left[\log \left(1 - GMIM\left(o, y\right)\right)\right].  
\end{split}
\end{equation}
\subsection{Loss Function}
Combining the cross-entropy loss with the LMIM and GMIM optimization function, the final objective loss function for the whole model is:
\begin{equation}
L_{total} = -\sum_{i=1}^{C} Y_{i} \log \hat{Y}_{i} - L_{(LMIM)} - L_{(GMIM)},
\end{equation}
\noindent where the first term is the cross-entropy loss and $ Y_{i}$ is the label. Because the three items in the above equation have the similar numbers in our experiments, we did not allocate different weights to each loss item in our implementation.
\section{Experiments}
In this section, we first evaluate the performance of our base architecture (\textbf{baseline}) which can be trained easier than previous methods. Then we conduct a thorough ablation study to the proposed LMIM and GMIM (GLMIM) and figure out how they help the model get better results respectively. we also compare with other state-of-the-art lip reading methods on two large word-level benchmarkss. Finally, we visualize the discriminative representations leaned with the GLMIM. Codes will be available at {\url{https://github.com/xing96/MIM-lipreading}}.
\subsection{Datasets}\label{4.1}
We evaluate our method on two large-scale word-level lip reading benchmarks, LRW and LRW-1000. The samples in both of these two datasets are collected from TV shows, with a wide coverage of the speaking conditions including the lighting conditions, resolution, pose, gender, make-up etc.

\textbf{LRW} \cite{chung2016lip}: It is released in 2016, including 500 word classes with more than a thousand speakers. It displays substantial diversities in the speaking conditions. The number of instances in the training set reaches 488766, and the number in validation and test set contains 25000 instances for each. LRW remains a challenging dataset and has been widely used by most existing lip reading methods.

\textbf{LRW-1000} \cite{yang2018lrw}: The dataset is a large-scale naturally distributed word-level benchmark, which has 1000 word classes in total. There are more than 70,000 sample instances in total, with a duration of about 57 hours. This dataset aims at covering a natural variability over different speech modes and imaging conditions to incorporate challenges encountered in practical applications. So the samples of the same word are not limited to a pre-specified length range, to allow the existence of various speech rates, which is consistent with the practical case and also brings more challenges. 
\subsection{Implementation Details}
The input frames in our implementation are all cropped or resized to $88\times88$ (Each video in LRW contains full face and the resolution is larger than $88\times88$, we cropped the mouth region by $88\times88$ directly; LRW1000 only contains the mouth region but the resolution is not fixed, we resized them to $88\times88$). The kernel size, stride and padding of the first 3D-CNN are $(5,7,7)$, $(1,2,2)$ and $(2,3,3)$ respectively. Each GRU or LSTM layer has 1024 hidden units (which means each Bi-GRU contains 2048 neurons). The Adam optimizer is applied for fast convergence. In the training process, the learning rate would decay from 0.0001 to 0.00001 when the accuracy doesn't increase. Dropout is utilized at each Bi-GRU layer to mitigate the overfitting problem.
\begin{table}[b]
\caption{Comparison of the modified baseline.}
\label{table1}
\begin{center}
\begin{tabular}{|c||c|}
\hline
Method & Accuracy\\
\hline
Petridis\cite{petridis2018end} & 82.00\% \\
\hline
Petridis\cite{petridis2018end}(our re-implement) & 81.70\% \\
\hline\hline
The Modified Baseline Architecture & \textbf{82.14\%}\\
\hline
\end{tabular}
\end{center}
\end{table}
\subsection{Baseline}
We adopt \cite{petridis2018end} as the base architecture. The accuracy of our re-implementation on LRW is a little lower than the value in the original paper. So we use the modified network as described in \ref{3.1} and take it as our \textbf{baseline} when using no MI constraint. Unlike \cite{petridis2018end}, we introduce the GAP layer to the modified network in order to get rid of training the front-end and the back-end separately. As shown in Table \ref{table1}, our modified architecture is superior to the base architecture, which achieves an accuracy of 82.14\% on the LRW dataset.
\subsection{Effect of the LMIM}\label{effort_LMIM}
In order to evaluate the effectiveness of the proposed LMIM, we train the baseline network with and without the LMIM separately. In both the two cases, the LMIM will be dropped when coming to test, which means that these two networks are totally the same in the test process. When we compare the accuracy between these two networks, we find that the network trained with the LMIM performs better. Besides the total accuracy, we conduct a further statistics analysis of the accuracy over each class. As shown in Table \ref{table_add1},  most classes with the LMIM show a higher accuracy and a clear improvement over the words with similar spellings or pronunciations, such as MAKES/MAKING and POLITICAL/POLITICIANS. This result shows that the proposed LMIM enable to extract the local fine-grained features indeed, which is significant to improve the ability to distinguish the words with similar pronunciations.
\begin{table}[t]
\caption{Examples of the improvement over words with similar pronunciations.}
\label{table_add1}
\begin{center}
\begin{tabular}{|c|c|c|c|}
\hline
Class & Baseline & Baseline with LMIM &  Improvement\\
\hline
\hline
MAKES & 62\% & 74\% & \textbf{12\%}\\
\hline
MAKING & 80\% & 92\% & \textbf{12\%}\\
\hline
\hline
POLITICAL & 82\% & 90\% & \textbf{8\%}\\
\hline
POLITICS & 84\% & 92\% & \textbf{8\%}\\
\hline
\hline
STAND & 48\% & 60\% & \textbf{12\%}\\
\hline
STAGE & 70\% & 80\% & \textbf{10\%}\\
\hline
\hline
NORTH & 78\% & 90\% & \textbf{12\%}\\
\hline
NOTHING & 78\% & 86\% & \textbf{8\%}\\
\hline
\hline
SPEND & 36\% & 46\% & \textbf{10\%}\\
\hline
SPENDING & 78\% & 82\% & \textbf{4\%}\\
\hline

\end{tabular}
\end{center}
\end{table}
\begin{figure}[b]
      \centering
      \includegraphics[width=0.9\linewidth]{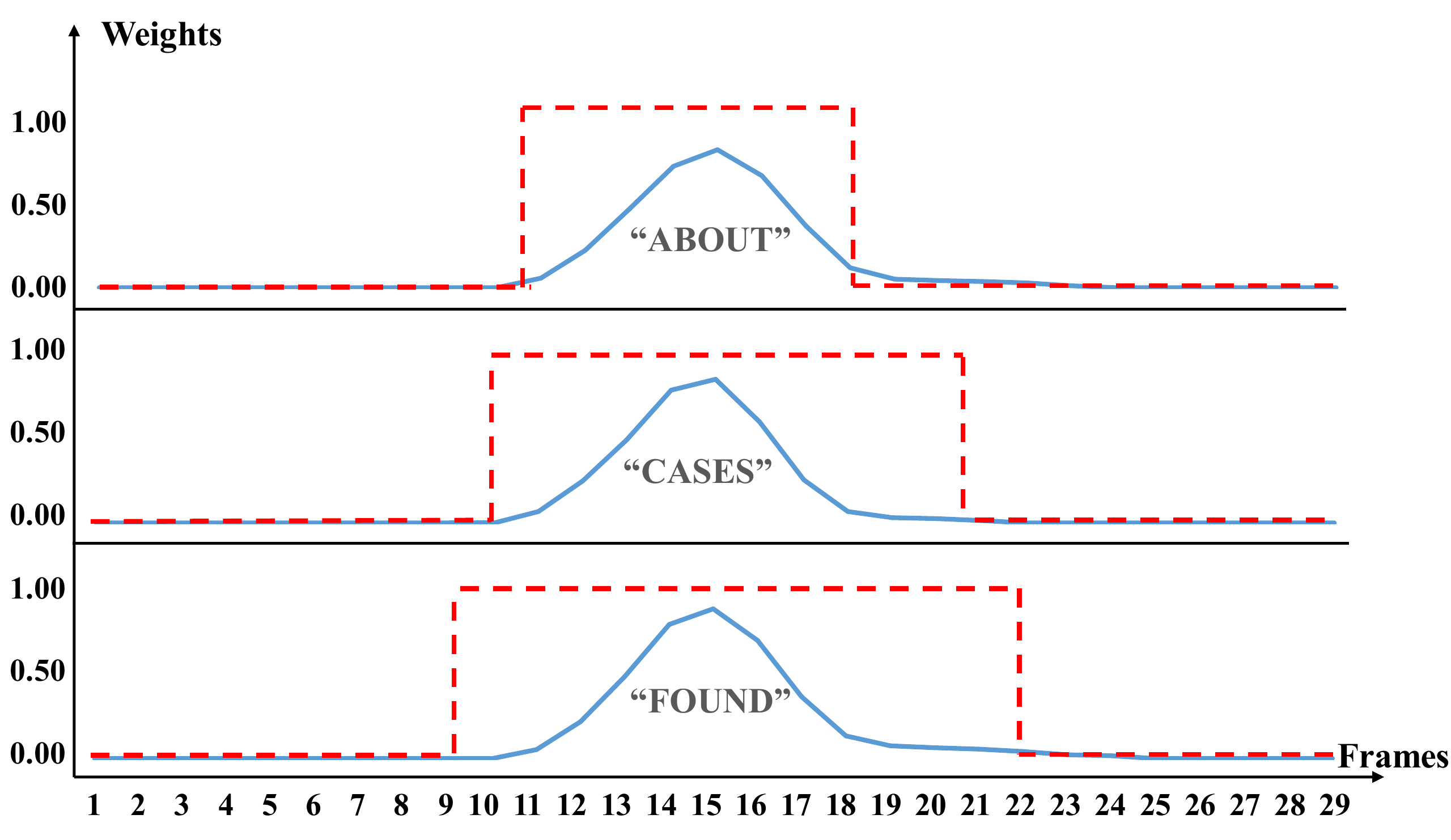}
       \caption{We randomly sample three words and show the weights of each frame learned with GMIM. The blue line shows the learned weight for each frame. The red dashed line denotes the word boundary for the target word when its value is 1.}
    \label{fig:4b}
\end{figure}
\subsection{Effect of the GMIM}
The ability to Select key frames is essential for lip reading because a video is always hard to cut to exactly containing only one word. This is why we apply GMIM to make the model pay different attention to all frames to select valid key frames. We directly based the experiments in this part on the model trained with LMIM in \ref{effort_LMIM} because of its excellent ability to extract fine-grained features. For the sake of fairness, the Front-end is fixed and only the Back-end is trained with GMIM. Without sending any additional word boundary information, we observed that the model has learned the key frames precisely and the accuracy has increased further. When the Front-end is trained together with the Back-end, we get a new state-of-the-art result.

The result of the weights $\beta$ learned with the proposed GMIM, is shown in Fig. \ref{fig:4b}. The horizontal axis represents the temporal dimension of the video, corresponding to 29 frames in the video. The vertical axis represents the numeric of the learned weights. The blue line shows the curve composed by the learned weights for each frame. The red dashed line with value 1 denotes the range divided by the annotated word boundary for the target word. Our model trained with GMIM not only learns the key frames successfully and pays more attention to the frames which are included in the word boundary, but also allocates small amount of weights to the frames close to the word boundary for capturing the context information. 

\subsection{Compare with state-of-the-art methods}
\begin{table}[t]
\setlength{\abovecaptionskip}{-0.2cm}   
\caption{Comaprison with other related work on LRW.}
\label{table2}
\begin{center}
\begin{tabular}{|c||c|}
\hline
Method & Accuracy \\
\hline\hline
Chung[2018]\cite{chung2018learning} & 71.50\% \\
Chung[2017]\cite{chung2017lip} & 76.20\% \\
Petridis[2018]\cite{petridis2018end} & 82.00\% \\
Stafylakis[2017]\cite{stafylakis2017combining} & 83.00\% \\
Wang[2019]\cite{wang2019multi} & 83.34\% \\
\hline\hline
Baseline & 82.14\% \\
Baseline+LMIM & 83.33\% \\
The Proposed GLMIM & \textbf{84.41\%}\\
\hline
\end{tabular}
\end{center}
\end{table}
\begin{table}[t]
\setlength{\abovecaptionskip}{-0.2cm}   
\caption{Comaprison with other related work on LRW1000.}
\label{table3}
\begin{center}
\begin{tabular}{|c||c|}
\hline
Method & Accuracy \\
\hline\hline
LSTM-5 & 25.76\% \\
D3D[2018]\cite{yang2018lrw} & 34.76\% \\
3D+2D & 38.19\%\\
Wang[2019]\cite{wang2019multi} & 36.91\% \\
\hline\hline
Baseline & \textbf{38.35\%} \\
Baseline+LMIM & \textbf{38.69\%} \\
The Proposed GLMIM & \textbf{38.79\%}\\
\hline
\end{tabular}
\end{center}
\end{table}
In this part, we compare the proposed GLMIM with the current state-of-the-art methods on both the two challenging benchmarks, LRW and LRW-1000. On the LRW dataset, although our baseline is not the best, the accuracy is improved for about 1.21\% after introducing the LMIM, which is expected to capture more discriminative and fine-grained features for the main task. Meanwhile, the GMIM improves the accuracy to 84.41\% furthermore, mainly beneficial from its advantage to pay different attention to different frames. Comparing with other lip reading methods which also have no extra inputs except the visual information, as shown in Table \ref{table2}, we get the best result and provide a new state-of-the-art result on the LRW dataset.

LRW1000 is another challenging large-scale benchmark, with a large variation of speech conditions including lighting conditions, resolution, speaker’s age, pose, gender, and make-up, etc. The best result is only 38.19\% up to now. It is challenging to obtain a good performance on this dataset while we achieve a high accuracy of 38.79\% which outperforms the existing state-of-the-art results. Table \ref{table3} gives the accuracy of our models. The improvement of the GMIM is smaller when comparing with the improvement on LRW, this interesting phenomenon may be due to that the number of useless frames in each word sample in LRW-1000 is smaller than LRW, which reduces the role of selecting key frames for each word.
\subsection{Visualization}


 \begin{figure*}[t]
\centering
\subfigure[Results before applying the GLMIM]{\includegraphics[width=0.4\linewidth]{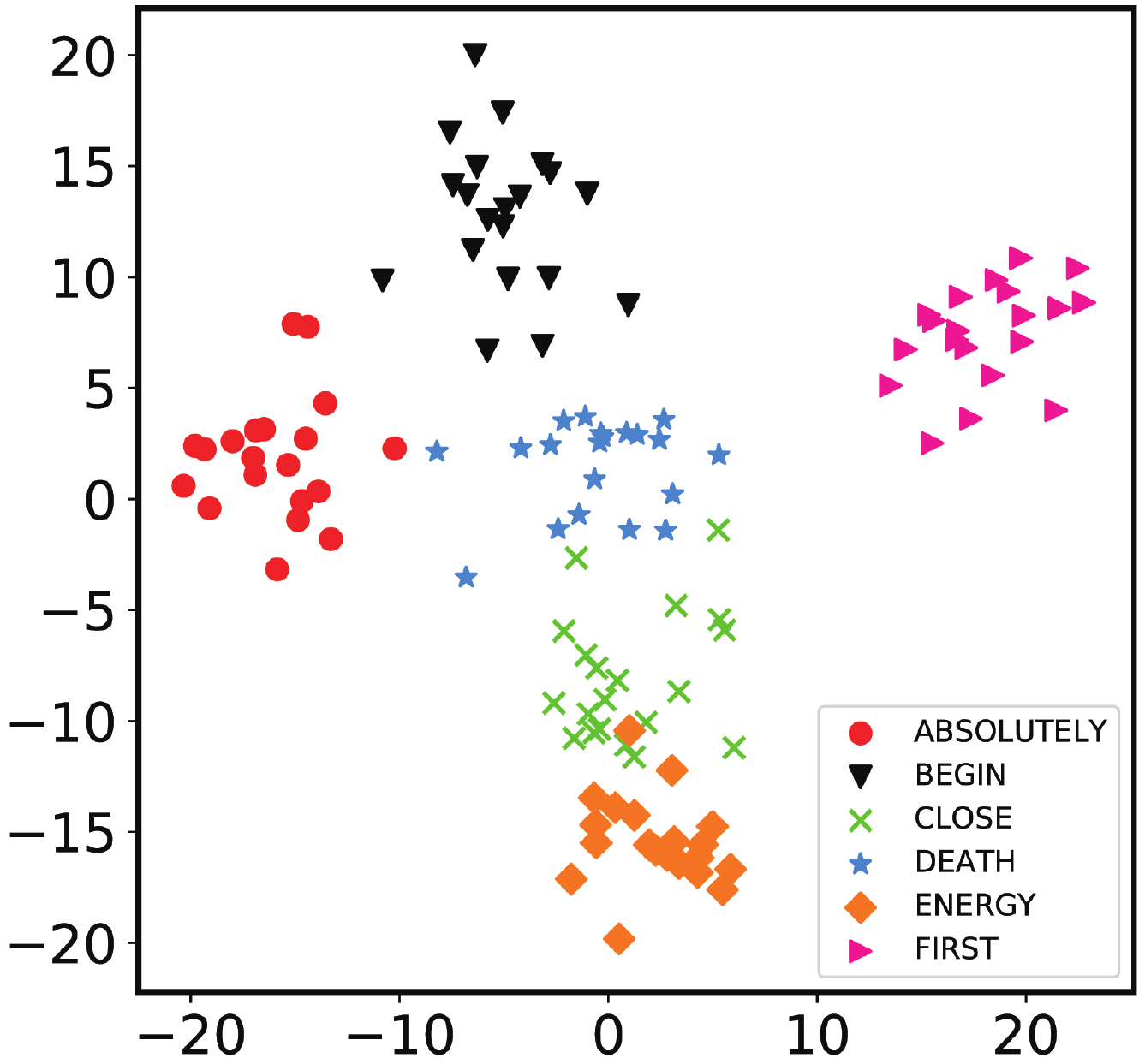}}
\hspace{0.6in}
\subfigure[Results after applying the GLMIM]{\includegraphics[width=0.4\linewidth]{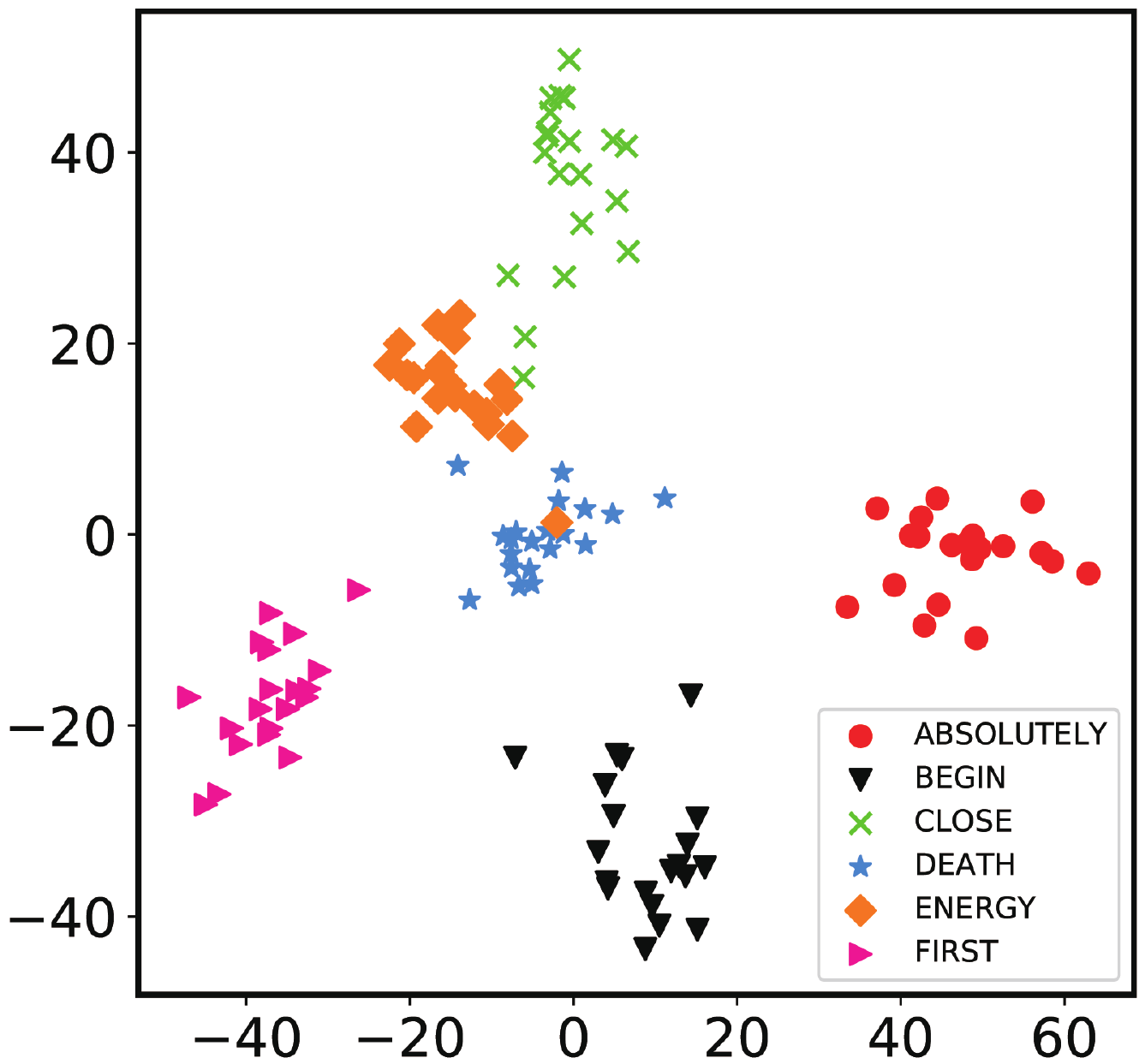}}
\caption{An example of the visualization for the final representations form the Bi-GRU. With the help of the GLMIM, the architecture gets more discriminative results.}
\label{fig5} 
\end{figure*}
In this section, we explore further the effect of the proposed GLMIM by visualization. Specifically, we randomly choose 6 classes and each of them contains 20 samples. We send them to thee model of our original baseline architecture with and without the proposed GLMIM respectively. Then we extract the final representations $\textbf{O}$ which will be sent to the linear layer for classification. We apply PCA to reduce its dimension form higher dimensions to 2 dimensions for better visualization. As is shown in Fig. \ref{fig5}, the variance among these classes before applying GLMIM ranges only from $-20$ to $20$; While the variance has been enlarged to the interval between $-40$ and $60$ after applying GLMIM, which means the variance among the classes have been greatly increased due to the introduction of the proposed GLMIM, which makes it easier to distinguish different classes.

\section{Conclusion}
 In this paper, we propose a mutual information maximization based method for both the local fine-grained feature extraction and global key frames selection. We also modify the existing model for lip reading that make it can be trained easier. We performed a detailed ablation study and obtain the best results on both the two largest word-level lip reading datasets.
\section{Acknowledgments}
This work is partially supported by National Key R\&D Program of China (No. 2017YFA0700804) and National Natural Science Foundation of China (No. 61702486, 61876171).

\bibliography{egbib}{}
\bibliographystyle{ieee}

\end{document}